\documentclass[twoside,11pt]{article}

\usepackage{caption}
\usepackage{subcaption}
\usepackage{tabularx} 
\usepackage{booktabs} 
\usepackage{array}    
\usepackage{jmlr2e}
\usepackage{amsmath} 
\usepackage{float}

\usepackage{lastpage}
\jmlrheading{23}{2022}{1-\pageref{LastPage}}{1/21; Revised 5/22}{9/22}{21-0000}{Author One and Author Two}

\firstpageno{1}
\begin{document}

\title{Self-supervised large-scale kidney abnormality detection in drug safety assessment studies}

\author{\name Ivan Slootweg \textsuperscript{1} \email ivan.slootweg@radboudumc.nl 
       \AND
       \name Natalia P. García-De-La-Puente \textsuperscript{2}  \email napegar@upv.es  
       \AND
       \name Geert Litjens \textsuperscript{1,3}*  \email geert.litjens@radbouudmc.nl 
       \AND
       \name Salma Dammak \textsuperscript{1,3,}*  \email salma.dammak@radbouudmc.nl \\
       \textsuperscript{1} \addr Department of Pathology, Radboud University Medical Center, Nijmegen, The Netherlands\\
       \textsuperscript{2} \addr Instituto Universitario de Investigación en Tecnología Centrada en el Ser Humano, Universitat Politècnica de València, Valencia, Spain\\
       \textsuperscript{3} \addr On behalf of the Bigpicture Consortium, \hyperlink{http://www.bigpicture.eu/}{https://www.bigpicture.eu/}\\ 
       \textnormal{*} Co-senior authors
       }

\editor{}

\maketitle

\begin{abstract}%

\end{abstract}
\noindent Kidney abnormality detection is required for all preclinical drug development. 
It involves a time-consuming and costly examination of hundreds to thousands of whole-slide images per drug safety study, most of which are normal, to detect any subtle changes indicating toxic effects. 
In this study, we present the first large-scale self-supervised abnormality detection model for kidney toxicologic pathology, spanning drug safety assessment studies from 158 compounds. 
We explore the complexity of kidney abnormality detection on this scale using features extracted from the UNI foundation model (FM) and show that a simple k-nearest neighbor classifier on these features performs at chance, demonstrating that the FM-generated features alone are insufficient for detecting abnormalities. 
We then demonstrate that a self-supervised method applied to the same features can achieve better-than-chance performance, with an area under the receiver operating characteristic curve of 0.62 and a negative predictive value of 89\%.
With further development, such a model can be used to rule out normal slides in drug safety assessment studies, reducing the costs and time associated with drug development.

\begin{keywords}
self-supervised learning, abnormality detection, toxicologic pathology
\end{keywords}

\section{Introduction}

Toxicologic pathology is the study of the effects of compounds on biological beings, which includes pre-clinical studies that assess the safety and effectiveness of novel therapies during drug development, before clinical trials.
Current regulations require that a novel drug be administered to experimental animals that are subsequently sacrificed so that their tissues can be comprehensively examined for toxicologic effects.
In each of these studies, tens to hundreds of experimental animals are used, resulting in 4,800 to 6,000 histology slides being produced per study \citep{mehrvar2021deep}.
Furthermore, several of these studies are necessary to test different dimensions of toxicity, including both short-term and long-term effects.
All this results in a very large number of tissue slides that require highly time-consuming and costly examination.
Therefore, if a model could automatically detect tissue abnormalities in these studies, it could significantly reduce the time and cost associated with safety testing during drug development, allowing unsafe drugs to be screened out more quickly, and safe, much-needed drugs, to reach patients faster.

To address this, several studies in the field have developed abnormality detection models, most of which focus on a single abnormality or a subset of abnormalities that are typically related to a specific toxicity mechanism.
However, some studies, summarized in Table \ref{tab:other-work}, also propose general abnormality models, which aim to detect any deviation from normal or healthy tissue morphology.
Such models are the most difficult to develop, as they target a very wide distribution of morphologies, but that also makes them the most useful in a drug-development setting.
\begin{table}[b]
\centering
\begin{tabular}{l c c c c c}
\hline
\\[-1em]

Study & Organ(s) & \parbox{2cm}{\centering Total WSIs} & \parbox{2.5cm}{\centering Metric} & \parbox{2.2cm}{\centering Metric value}\\
\\[-1em]
\hline
\\[-1em]
\cite{kuklyte2021evaluation} & 5 organs\textsuperscript{$\dagger$} & 1,342 & Pixel-level F1-score & 0.63--0.84\textsuperscript{$\ddagger$}  \\
\cite{zingman2024learning} & Liver & 700 & Balanced accuracy & 0.96 \\
\cite{su2023prediction} & Liver & 612 & AUC & 0.84 \\
\cite{freyre2021biomarker} & Kidney & 349 & AUC & 0.97 \\
\cite{shelton2024automated} & 39 organs\textsuperscript{$\dagger$} & 54 & Patch-level AUC & 0.94 \\
\\[-1em]
\hline
\multicolumn{5}{l}{\textsuperscript{$\dagger$}Both multi-organ studies include kidney and liver.}\\
\multicolumn{5}{l}{\textsuperscript{$\ddagger$}The range is across organs. The highest F1-score for the kidneys is 0.84.}
\end{tabular}
\caption{Studies proposing general abnormality detection models.}
\label{tab:other-work}
\end{table}

Using supervised approaches with detailed pixel annotations, \cite{kuklyte2021evaluation} train and test various pipelines to segment abnormalities.
\cite{zingman2024learning} and \cite{freyre2021biomarker} use supervised training to build whole-slide image (WSI) level abnormality detection models. 
They first extract patch-level features using models pre-trained on an auxiliary task (e.g., organ detection), then they use different methods and models to aggregate them into WSI-level features.
This is followed by training a simple classifier (e.g., a single neuron) to distinguish WSIs with abnormalities.
\cite{su2023prediction} also use a pre-trained network for patch feature extraction, but then use a graph neural network approach to classify the WSIs.
All these supervised models have strong performances, as shown in Table \ref{tab:other-work}.

These studies show that a supervised approach can perform well for abnormality detection, but supervised training for this problem carries a critical limitation.
Drug safety assessment studies can exhibit a broad distribution of abnormalities, and when a new compound is being tested, unexpected and previously unseen abnormalities could occur.
This makes it impossible to comprehensively represent the abnormal class when building a supervised abnormality detection model.
In contrast, the category of normal tissue is well-defined and forms a consistent class.
Self-supervised methods can learn exclusively from normal tissue data, and therefore could work well for this problem \citep{SSAdam}. 

\cite{shelton2024automated} propose such a model.
They use a ResNet-based encoder from \cite{he2016residual} to represent each patch in a latent space which they use as input to a generative adversarial network (GAN) that then reconstructs the patches.
They only train their model on normal tissue patches, so when abnormal patches are presented at testing, their reconstruction is poorer than that of normal patches.
This difference in reconstruction score allows them to detect abnormal morphologies with a patch AUC of 0.94, when testing on patches from 14 WSIs.
While this is a small study, it shows that a self-supervised approach could allow for general abnormality detection.

\begin{figure}
    \centering
    \includegraphics[width=0.9\linewidth]{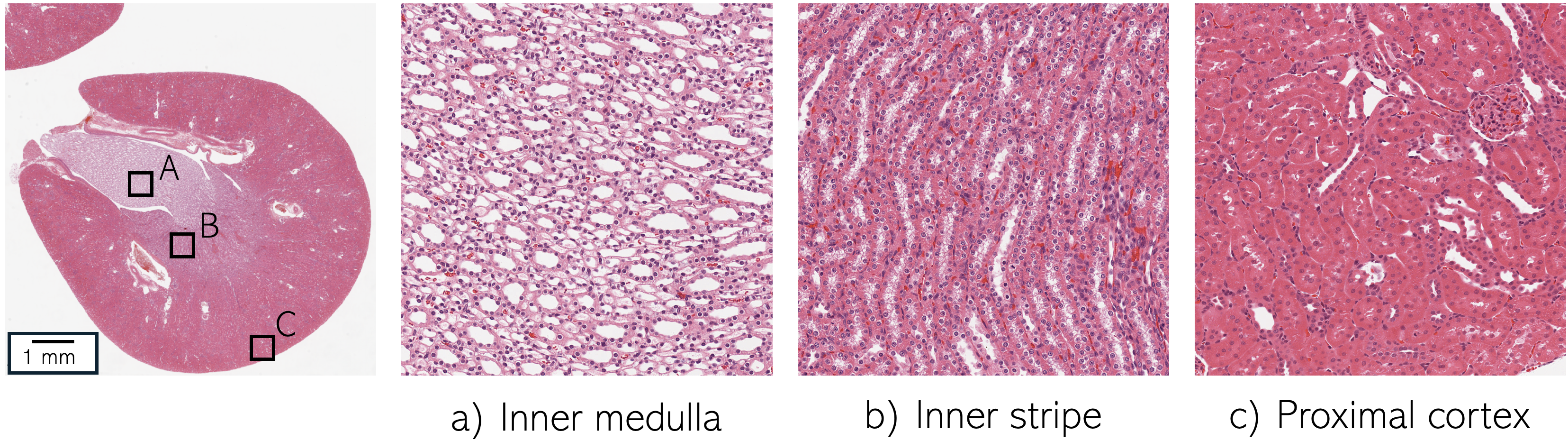}
    \caption{A kidney with no abnormalities, images a-c are 500$\mu$m$\times$500$\mu$m.}
    \label{fig:normal_kidney}
\end{figure}

In our study, we investigate whether a self-supervised model can detect abnormalities at the WSI level in a large multi-study setting.
We focus our investigation on the kidney, as it is one of the primary sites for toxin filtration and metabolism, and as it contains various normal morphologies, shown in Fig. \ref{fig:normal_kidney}, which makes detecting abnormalities in it more challenging.
We leverage a digital pathology foundation model (FM) for feature extraction and test whether these feature vectors are sufficient for abnormality detection with a simple classifier.
Then we develop a self-supervised autoencoder for reconstructing these feature vectors, and test whether using that improves performance.

\section{Methods}

\subsection{Dataset}

For this study, we used the 29,168 kidney WSIs provided in the Open TG-GATES database \citep{igarashi2015open}.
This is a publicly available database containing kidney and liver WSIs, diagnoses, and genomic data from drug safety assessment studies on male Sprague-Dawley rats that were collected by the Japanese Toxicogenomics Project consortium.
The WSIs in this database originate from studies of 158 different compounds, which exhibit vastly different mechanisms of action and cause various abnormality rates, types, and severity levels.
Subsequently, the WSIs contain differences in tissue morphology that vary from study to study.
Additionally, the drug safety assessment studies were conducted by three different Contract Research Organizations (CROs) with different equipment and protocols, and the database was compiled over a 10-year period. 
Due to this, the WSIs likely accumulated additional variation.
All slides were stained with hematoxylin and eosin and scanned using the ScanScope AT (Leica Biosystems, Nussloch, Germany) at 0.49 ${\mu}$m/pixel (20$\times$).

\begin{figure}[h!]
    \centering
    \includegraphics[width=1\linewidth]{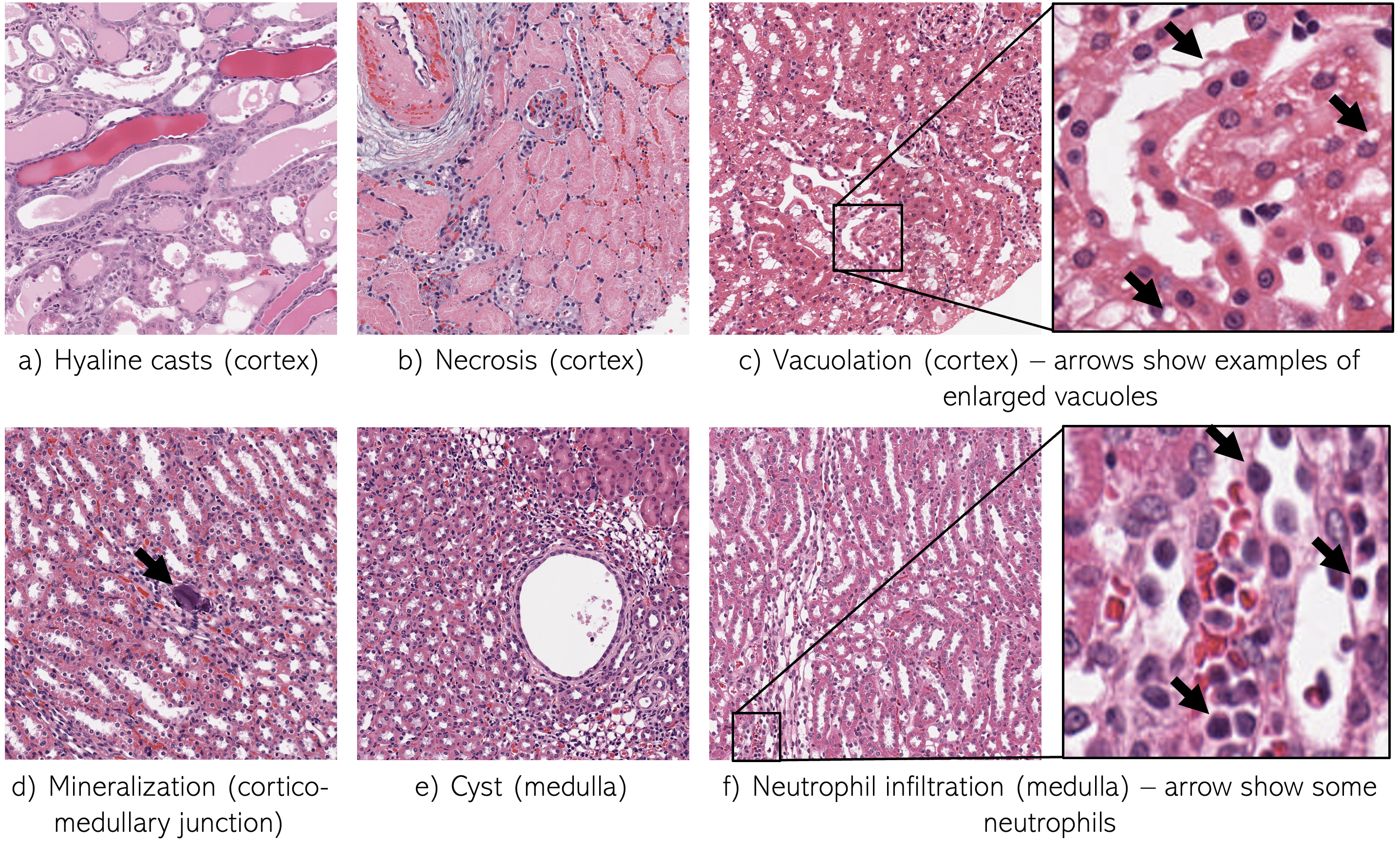}
    \caption{Examples of abnormalities with kidney sub-regions in brackets. All main images are 500 $\mu$m$ \times$ 500 $\mu$m. The zoomed in image in c is 100 $\mu$m $\times$ 100 $\mu$m. The zoomed in image in f is 68 $\mu$m $\times $68 $\mu$m}
    \label{fig:abnormalities}
\end{figure}

The CROs first collected the abnormality labels, then the pathologists from the consortium companies peer-reviewed and harmonized them.
These labels were recorded per animal.
As animals could have multiple slides, usually 2-3 if an abnormality is detected, and the label could apply to any number of the animal's slides, we applied the diagnosis label to each WSI taken from that animal.
These labels indicate whether an abnormality is present (i.e., normal vs. abnormal) and specify its type among 56 possibilities, a few of which are shown in Figure \ref{fig:abnormalities}.

\cite{kuklyte2021evaluation} has shown that the scale of the abnormality has a large effect on model performance.
To explore this, we grouped the 56 abnormalities in our dataset into five ordinal categories based on their scale.
In order of increasing scale, they are subcellular, cellular, multicellular, tissue-level/lesions, and unspecified.
Subcellular abnormalities include localized changes at the cell organelle level.
Cellular abnormalities are those where the whole cell deviates from normal morphology, such as immune infiltration.
Multicellularity involves multiple cells clustered together or spread throughout tissues. 
Tissue-level/lesions describe changes in tissue that are large and apparent at low magnification.
Lastly, the unspecified category includes systemic findings such as death.
The exact groupings are in Appendix \ref{sec:appendix} Table \ref{tab:scale_groups}.
The groups are not mutually exclusive, as a single slide may contain multiple abnormalities. 
We considered a slide to belong to a group if it exhibited any abnormality from that group, regardless of other co-occurring types.

We split the data by compound to test the robustness of our model on unseen test compounds, which represents the most realistic use-case scenario.
Furthermore, as drug safety assessment studies tend to have substantially more normal slides than abnormal ones, we aimed to preserve this imbalance in our test set.
However, directly matching the percentage of abnormalities across subsets was not possible as we chose to split the data by compound.
Instead, we constructed the subsets by adding compounds one by one, subsequently evaluating the label balance and choosing which subset to add it to based on the resulting label distribution.
We repeated this 1000 times and picked the most balanced dataset split among them.
We performed this in two steps: first, splitting the compounds into testing and non-testing sets, and subsequently splitting the non-testing compounds into training and validation sets. 
As the model is only trained on the normal slides, we excluded abnormal slides from training compounds to ensure that validation reflects the performance on unseen compounds. The dataset split results are summarized in Table \ref{tab:subsets}.

\begin{table}[h]
    \centering
    \begin{tabular}[h!]{ccccc}
        \hline
         Subset &  No. compounds & No. WSIs & No. animals & \begin{tabular}{@{}c@{}}No. WSIs with \\ abnormalities (\%)\end{tabular}\\
         \hline
         Training   &  83 &  12,341 &  9,949 & 0 (0\%)\\
         
         Validation &  41 &   7,292 &   5,961 & 1,125 (15\%)\\
         Testing    &  34 &   7,288 &   5,143 & 1,119 (15\%)\\
         \hline
    \end{tabular}
    \caption{Summary of the experimental subsets.}
    \label{tab:subsets}
\end{table}

\subsection{Feature space}

For feature extraction, we use the slide2vec WSI feature extraction library \citep{grisi2025}.
First, we use Otsu's thresholding at 5$\times$ magnification to segment tissue as foreground.
From within the detected tissue regions, we extract all $224\times224$ pixel patches at full resolution (0.49 mpp, 20$\times$). 
We then use the UNI FM to extract a 1024-dimensional feature embedding from each patch \citep{chen2024uni}. 
We chose UNI for feature extraction as it was trained on a variety of organs, with a large number of kidney slides (\textgreater 8,000 kidney WSIs), and a broad variety of diagnoses, including neoplastic, infectious, inflammatory, and normal morphologies.
Most importantly, UNI was trained entirely on non-public data, ensuring that we do not accidentally test it on WSIs it was trained on, given that we use public data for this study.

To obtain WSI-level embeddings, we aggregate patch features using two different methods: mean pooling and max pooling.
To explore the information of the UNI features we generate, we use t-SNE to project the high-dimensional feature vectors to 2-dimensions, where each WSI is represented by a point in this feature space \citep{TSNE}. 
The number of compounds in this study makes it difficult to visualize compound-specific clusters in a way that is human-interpretable. 
Therefore, only for the plots in this paper, we selected the ten compounds with the largest numbers of WSIs to visualize.
When assessing whether WSIs with the same label cluster together, we use all WSIs in the t-SNE plot.

To establish a baseline for determining whether the UNI features alone can distinguish abnormalities, we apply a k-nearest neighbor (kNN) algorithm with k=1 and uniform weights. 
The trained kNN holds the feature space of the normal training WSI's, and returns the distance of each test WSI to the nearest training sample.
We treat these distances as abnormality scores, with a greater distance signifying a greater likelihood of abnormality.
We test the kNN for both types of WSI-level aggregation.

\subsection{Self-supervised model}
We selected the architecture of the self-supervised model by comparing the performance of different configurations using the area under the receiver operating characteristic curve (AUC) on a subset of the training and validation sets.
In this architecture optimization step, we compared mean and max pooling for obtaining WSI-level features, the use of attention, and the use of either a sparse autoencoder (SAE) or a residual variational autoencoder, as well as different layer configurations. For the architecture we selected the SAE with two linear layers ($512$, $256$) in the encoder and decoder, along with squeeze-and-excitation channel attention \citep{8578843}. Attention is applied prior to the encoder to focus reconstruction on meaningful UNI features. 
Sparsity plays a crucial role in enhancing disease-related information. 
SAEs \citep{Makhzani2013kSparseA}  enforce a compact, disentangled representation of concepts by restricting activations to a small subset of neurons via L1 regularization on the weight values:

\vspace{-0.5em}
\[
\mathcal{L}_{\text{SAE}} = \text{MSE}(x, \hat{x}) + \lambda \|z\|_1.
\]
where \(x\) is the input, \(\hat{x}\) the reconstruction, \(z\) the latent vector, and \(\lambda\) the sparsity
weight which was set to $1$. 

During training, the autoencoder learns to reconstruct feature vectors in WSIs without abnormalities from compact representations generated by the encoder.  Since we only present normal samples to the model during training, it is expected to reconstruct unseen abnormal data poorly. To apply the autoencoder for distinguishing normal from abnormal slides, we represent the likelihood of abnormality by the mean-squared reconstruction error (MSE). 
We train the model for 50 epochs with a learning rate of $ 1 \times 10^ {-5}$, which decreases by a factor of $0.1$ every five epochs when the MSE on the validation set has not improved. 
The model from the epoch with the lowest validation MSE is used for inference. 
In inference, MSE is used as a score for abnormality detection. 
For evaluation, we calculate the AUC and use Youden’s J-index on the validation set to determine an operating point, which enables us to calculate the sensitivity, specificity, positive predictive value (PPV), and negative predictive value (NPV) for both the validation and test sets.
To gain a better understanding of model performance, we calculate the model's disaggregated sensitivity for each category of the abnormality scale.

\section{Results \& Discussion}

\subsection{Feature space} 
\begin{figure}[htbp]
\centering

\begin{subfigure}{0.45\textwidth}
    \centering
    \includegraphics[width=\textwidth]{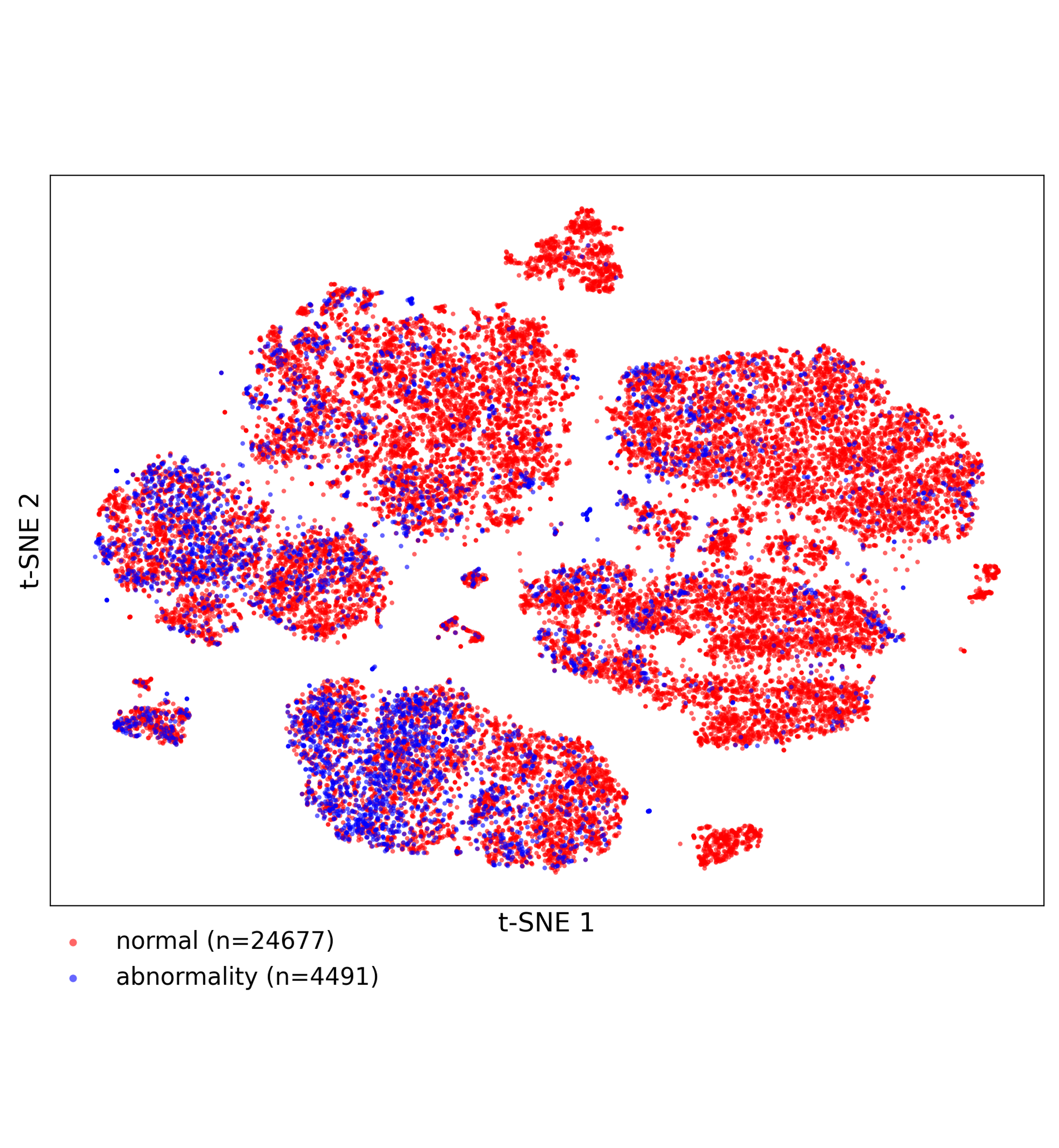}
    \caption{WSI features labeled by abnormality from all kidney WSIs in the Open TG-GATES}
    \label{fig:tsne_mean_study}
\end{subfigure}
\hfill
\begin{subfigure}{0.45\textwidth}
    \centering
    \includegraphics[width=\textwidth]{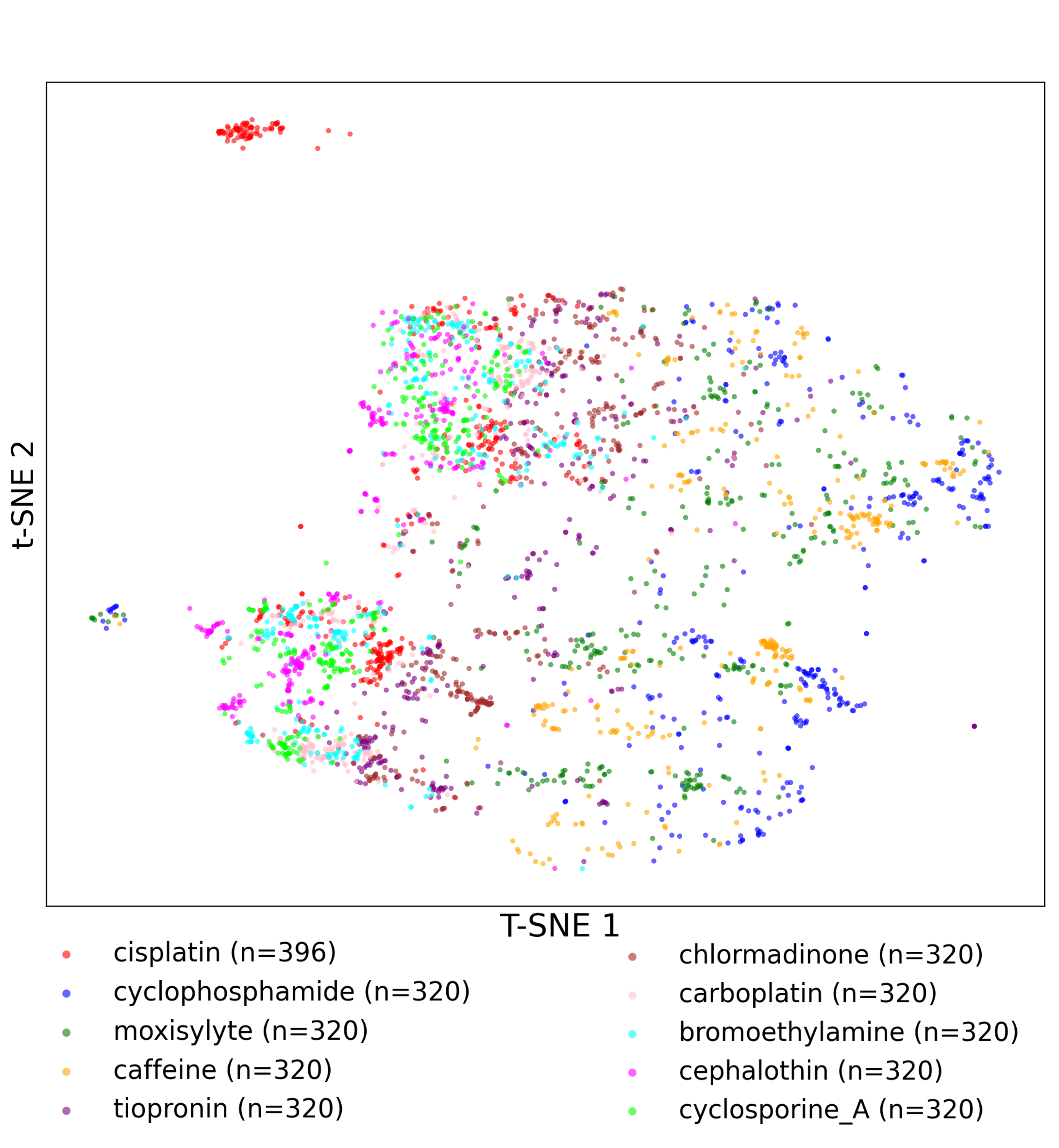}
    \caption{WSI features labeled by compound for ten compounds with the most WSIs}
    \label{fig:tsne_max_study}
\end{subfigure}
\caption{2D t-SNE projections of WSI feature vectors from max-pooling patch features.}
\label{fig:tsne_pooled_projections}
\end{figure}

Figure \ref{fig:tsne_pooled_projections} displays the max-pooled WSI-level UNI feature space, illustrating the complexity of our dataset.
As shown in Figure \ref{fig:tsne_pooled_projections}a, in the low-dimensional t-SNE space, abnormal WSI embeddings are not easily separable from normal ones.
We observed the same pattern with mean-pooling.
This was further reflected in the kNN classifier's performance applied in the full high-dimensional space.
The kNN classifier had a test set AUC of 0.50 for max pooling and 0.48 for mean pooling.
The validation set AUCs were the same.

Additionally, we do observe clustering by compound.
Without this clustering, labels in Figure \ref{fig:tsne_pooled_projections}b would be more uniformly distributed.

While the figure shows only the ten compounds with the most WSIs, similar clustering is seen across the remaining compounds and with mean pooling.
However, t-SNE is only a projection and no firm conclusions can be drawn.

\subsection{Self-supervised model}
The proposed self-supervised model had an AUC of 0.62 on the test set as shown in Table \ref{tab:val_test_performance}, demonstrating that a model can distinguish abnormal from normal kidney slides without any training with abnormal WSIs.
This indicates that while the feature space was potentially too noisy to separate the two classes using a simple kNN, it was sufficiently representative for reconstructing features from normal slides compared to abnormal ones, enabling indirect separation between the two classes.
This suggests that an unsupervised reconstruction loss approach may be generally practical even when the feature space does not naturally separate target labels.
Upon examining additional error metrics, we observe that the model achieves a balanced sensitivity and specificity.
This means that it is about as effective for detecting abnormal slides as it is for detecting normal slides.
While neither value is particularly high, this balance confirms that this approach does not simply label all slides as normal, despite abnormal slides being a lot more common.
We do, however, see a marked imbalance in NPV and PPV.
When the model predicts that a WSI is negative, it is much more likely to be accurate than when it predicts a sample as positive.
This could be due to unexpected differences on normal slides, such as artifacts or ink of an uncommonly used color.
In such an example, the model is likely to have a poor reconstruction score for the slide, despite no pathological abnormality being present, and thus label it as abnormal.  However, that remains to be tested. Regardless, with an NPV of 89\%, the model could be used as a triaging tool, ruling out a large number of slides with no abnormalities, leaving a much smaller subset for the pathologist to examine first.
\begin{table}[t]
    \centering
    \begin{tabular}{lcc}
        \hline
        Metric & Validation Set & Test Set \\
        \hline
        AUC            & 0.63 & 0.62 \\
        Sensitivity    & 80\% & 62\% \\
        Specificity    & 42\% & 58\% \\
        PPV            & 43\% & 21\% \\
        NPV            & 79\% & 89\% \\
        \hline
    \end{tabular}
    \caption{Performance of the self-supervised model on the validation and test sets}
    \label{tab:val_test_performance}
\end{table}

When examining model sensitivity for the abnormality categories separately, we see an increase in sensitivity that is consistent with the increase in abnormality scale. This means that the model is better able to detect abnormalities the larger they are, which makes intuitive sense, as larger abnormalities have more representative pixels and hence more information for the model to use. This is consistent with the findings in \cite{kuklyte2021evaluation}.
We also see that there is no apparent relationship between the number of samples and performance in each of these categories, indicating that this effect is independent of sample size.
We also found that the different scales do not form clusters in a t-SNE projection of the feature space as shown in Appendix \ref{sec:appendix} Fig. \ref{fig:abnormality_scale}.

\begin{table}
    \centering
    \begin{tabular}{lccc}
        \hline
        Abnormality scale category &  Sensitivity & No. test samples \\
        \hline
        Subcellular &  54\% & 514 \\
        Cellular &  64\% & 309 \\
        Multi-cellular & 63\% & 390 \\
        Tissue-level/lesions & 67\% & 63 \\
        Unspecified & 79\% & 64 \\
        \hline
    \end{tabular}
    \caption{Self-supervised model performance sorted by abnormality severity.}
    \label{tab:self_supervised_sorted}
\end{table}
\subsection{Limitations and future work}

The performance of our model, at an AUC of  $0.62$,  is below that of previously published models for general abnormality detection, which range in AUC/F1/balanced accuracy between $0.63$ and $0.97$, as shown in Table \ref{tab:other-work}.
However, excluding \cite{shelton2024automated}, all other models are supervised for abnormality detection, which makes the task less challenging due to the availability of labels.
The self-supervised model of \cite{shelton2024automated}  exceeds the performance of ours, at an AUC of 0.94, but this performance is on a patch level, rather than a WSI level, making it not directly comparable.
Furthermore, all these studies were tested on datasets that were substantially smaller than ours.
\cite{kuklyte2021evaluation}, the largest study among them, used 15\% of their dataset for testing, which is 201 WSIs, while our model was tested on 7,288 WSIs.
Regardless, to be effective for use in the lab, our model's performance likely needs to improve, and its design has several limitations that may have limited its performance.
Firstly, we assigned labels to each animal's WSIs. 
It would have been better to pool the UNI features from all patches of that animal's slides and label that bag of WSIs instead, as it is possible that an abnormality occurred in only one of the slides of that animal.
We also used a relatively simple thresholding algorithm for tissue segmentation.
While this approach performs reasonably well, it is not the best available option.
In the case of very small abnormalities, a missed tile could result in a missed diagnosis, and it is possible that this algorithm overlooked some of these key tiles.

Lastly, we used an SAE but more sophisticated self-supervised approaches exist, such as generative modeling. For example, GANs or diffusion models can detect abnormalities by learning the distribution of normal data.
For this task, \cite{shelton2024automated} have shown that GANs can produce a high performance.
For a clinical pathology task, \cite{Linmans2024} showed than diffusion models can achieve excellent out-of-distribution detection performance.
Given the large scale of this study, we did not have the computational resources to evaluate either, as training for both methods is highly computationally intensive and slow. 
However, we hope to evaluate these methods in future work.

\section{Conclusion}

In this study, we found that a self-supervised model can detect kidney slides with abnormalities with an AUC of 0.62 and an NPV of 89\% on a test set of 7,288 slides from safety studies of 34 different compounds.
With further development, such a model can be utilized during drug development to de-prioritize normal slides during drug safety assessment studies.
This could reduce the costs and time associated with drug development, ultimately leading to quicker detection of unsafe drugs and for much-needed drugs to reach patients faster.

\acks{This project has received funding from the Innovative Medicines Initiative 2 Joint Undertaking under grant agreement No 945358. This Joint Undertaking receives support from the European Union’s Horizon 2020 research and innovation program and EFPIA. \hyperlink{www.imi.europe.eu}{www.imi.europe.eu}. We would also like to thank Adrian Wolny and Santiago Villalba for their help with downloading the Open TG-GATES dataset.}
\newpage

\bibliography{sample}

\newpage
\appendix{}
\section{Supplementary materials}
\label{sec:appendix}

\setcounter{figure}{0}
\renewcommand{\thefigure}{S\arabic{figure}}
\setcounter{table}{0}
\renewcommand{\thetable}{S\arabic{table}}

\begin{figure}[h!]
    \centering
    \includegraphics[width=1\linewidth]{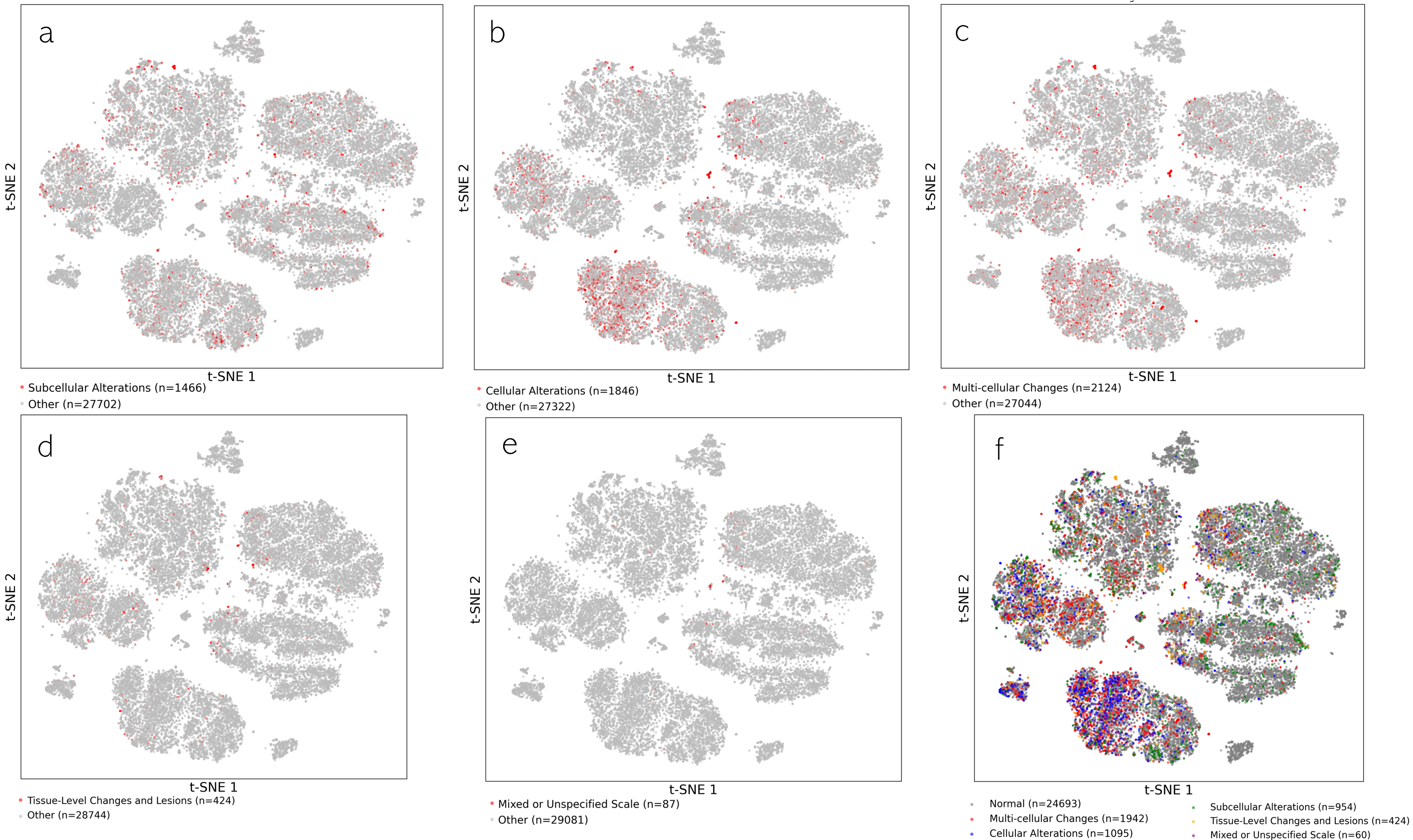}
    \caption{2D t-SNE projections of WSI UNI feature vectors from max-pooling patch features. For (a-e) we separate abnormalities by scale and indicate the WSIs for which any abnormality of a certain scale is present in red, all other WSIs are displayed in grey. These labels are not mutually-exclusive, as a single slide may contain multiple abnormalities. Therefore, in (f) we label the WSIs according to the largest scale of all abnormalities, if any, in a WSI. With these plots we show that UNI feature space is unlikely to separate abnormalities by their scale.  }
    \label{fig:abnormality_scale}
\end{figure}

\newcolumntype{L}[1]{>{\raggedright\arraybackslash}p{#1}}

\begin{table}[h!]
\centering
\renewcommand{\arraystretch}{1.3}
\begin{tabularx}{\textwidth}{L{3.5cm} L{4.5cm} X}
\toprule
\textbf{Category} & \textbf{Explanation} & \textbf{Included Abnormalities} \\
\midrule

\textbf{Subcellular Alterations} &
Localized changes at the cell organelle level. &
Karyomegaly; Vacuolization (Vacuolation): cytoplasmic; Eosinophilic body; Hyaline droplet; Calcification; Mineralization; Deposit: pigment; Inclusion body: intracytoplasmic; Change: basophilic; Alteration: nuclear; Alteration: cytoplasmic; Anisonucleosis \\

\addlinespace
\textbf{Cellular Alterations} &
Whole-cell scale deviations from normal physiology. &
Hyperplasia; Hyperplasia: regenerative; Hypertrophy; Hypoplasia; Degeneration; Degeneration: hydropic; Swelling; Desquamation; Increased mitosis; Regeneration; Dysplasia; Atypia: cellular \\

\addlinespace
\textbf{Multi-cellular Changes} &
Changes involving multiple cells in clusters or spread throughout the tissues. &
Fibrosis; Edema; Cyst; Cyst: hemorrhagic; Dilatation; Dilatation: cystic; Inflammation; Arteritis; Tubulitis; Cellular infiltration; Cellular infiltration: lymphocyte; Cellular infiltration: mononuclear cell; Cellular infiltration: neutrophil; Proliferation; Thickening; Cast: cellular; Cast: hemoglobinogenous; Cast: hyaline; Congestion; Sclerosis: glomerulus; Angiectasis; Arteriolosclerosis \\

\addlinespace
\textbf{Tissue-Level Changes and Lesions} &

Pathological findings which are large and apparent at low magnification.&
Necrosis; Infarct; Nephroblastoma; Hydronephrosis; Scar; Granuloma; Hemorrhage \\

\addlinespace
\textbf{Mixed or Unspecified Scale} &

Unspecified or non-specific categories.
&
Death; Lesion: NOS; Bacterium; Not specified \\

\bottomrule
\end{tabularx}
\caption{Biology-informed aggregation of abnormalities by scale to reflect the approximate biological scale and interpretability at which abnormalities become apparent, from fine-grained subcellular structures to broad tissue-level lesions.}
\label{tab:scale_groups}
\end{table}

\end{document}